\pdfoutput=1

\documentclass[11pt]{article}

\usepackage[preprint]{sty/acl}

\usepackage{times}
\usepackage{latexsym}

\usepackage[T1]{fontenc}

\usepackage[utf8]{inputenc}

\usepackage{microtype}

\usepackage{inconsolata}

\usepackage{graphicx}

\usepackage{booktabs}       
\usepackage{amsfonts}       
\usepackage{amssymb}
\usepackage{nicefrac}       
\usepackage{microtype}      
\usepackage{xcolor}         
\usepackage{amsmath}
\usepackage{amsfonts,amssymb}
\usepackage{multirow}
\usepackage{booktabs}
\usepackage{hyperref}
\usepackage{float}
\usepackage{enumitem}
\usepackage{algorithmicx}
\usepackage{algpseudocode}
\usepackage{tabularx}
\usepackage{subcaption}
\usepackage{marvosym}
\usepackage{adjustbox} 
\usepackage{textgreek}
\usepackage{wrapfig}
\usepackage{algorithm}
\usepackage{algpseudocode}
\usepackage{textcomp}
\usepackage{fontawesome}
\usepackage{xspace}
\usepackage[most]{tcolorbox}
\usepackage{colortbl}
\usepackage{url}   

\usepackage{amsmath,amsfonts,bm}

\def\eqref#1{equation~\ref{#1}}

\def\1{\bm{1}}

\DeclareMathAlphabet{\mathsfit}{\encodingdefault}{\sfdefault}{m}{sl}
\SetMathAlphabet{\mathsfit}{bold}{\encodingdefault}{\sfdefault}{bx}{n}

\definecolor{darkblue}{rgb}{0, 0, 0.5}
\hypersetup{citecolor=darkblue}
\definecolor{lightred}{RGB}{255,230,230}
\definecolor{lightgreen}{RGB}{230,255,230}
\definecolor{mediumgreen}{RGB}{200,255,200}
\definecolor{darkgreen}{RGB}{150,255,150}
\definecolor{deepdarkgreen}{RGB}{0,120,0}
\definecolor{lightred}{RGB}{255,230,230}
\definecolor{promptgreen}{RGB}{145, 204, 117}

\newtcolorbox{promptbox}[2][Prompt]{
colback=black!5!white,
arc=5pt, 
boxrule=0.5pt,
breakable,
enhanced,
fonttitle=\bfseries,
left=0pt,
right=0pt,
top=0pt,
bottom=0pt,
title=#1, 
before upper={\small}, fontupper=\fontfamily{ptm}\selectfont,
colframe=#2,
}
\newcommand{\ours}{\textsc{LongRePS}\xspace}

\title{Chain-of-Thought Matters: Improving Long-Context Language Models with Reasoning Path Supervision}

\author{%
  Dawei Zhu~\thanks{Equal contribution.}~$^{\heartsuit}$ \quad Xiyu Wei~\footnotemark[1]~$^{\heartsuit}$ \quad Guangxiang Zhao~\thanks{Primary mentor}~$^\diamondsuit$ \quad Wenhao Wu~$^{\heartsuit}$ \quad Haosheng Zou~$^{\diamondsuit}$ \\
  {\bf Junfeng Ran~$^{\heartsuit}$ \quad Xun Wang~$^{\heartsuit}$ \quad Lin Sun~\thanks{Corresponding authors.}~$^{\diamondsuit}$ \quad Xiangzheng Zhang~\footnotemark[3]~$^{\diamondsuit}$ \quad Sujian Li~\footnotemark[3]~$^{\heartsuit}$} \\[5pt]
  $^\heartsuit$~Peking University \quad $^\diamondsuit$~Qiyuan Tech \\
  \\
  {\texttt{\url{https://github.com/lemon-prog123/LongRePS}}}
}

\begin{document}
\maketitle

\begin{abstract}
Recent advances in Large Language Models (LLMs) have highlighted the challenge of handling long-context tasks, where models need to reason over extensive input contexts to aggregate target information. While Chain-of-Thought (CoT) prompting has shown promise for multi-step reasoning, its effectiveness for long-context scenarios remains underexplored. Through systematic investigation across diverse tasks, we demonstrate that CoT's benefits generalize across most long-context scenarios and amplify with increasing context length. Motivated by this critical observation, we propose \ours, a process-supervised framework that teaches models to generate high-quality reasoning paths for enhanced long-context performance. Our framework incorporates a self-sampling mechanism to bootstrap reasoning paths and a novel quality assessment protocol specifically designed for long-context scenarios. Experimental results on various long-context benchmarks demonstrate the effectiveness of our approach, achieving significant improvements over outcome supervision baselines on both in-domain tasks (+13.6/+3.8 points for LLaMA/Qwen on MuSiQue) and cross-domain generalization (+9.3/+8.1 points on average across diverse QA tasks). Our code, data and trained models are made public to facilitate future research.
\end{abstract}

\section{Introduction}

Large Language Models (LLMs) have revolutionized language modeling and achieved remarkable success in traditional NLP tasks~\citep{brown2020language}. This success has spurred their application to increasingly complex real-world scenarios~\citep{guo2025deepseek,claude3.5sonnet,GPT4o}, particularly long-context tasks such as document-level question answering~\citep{trivedi2022musique}, summarization~\citep{zhong2021qmsum}, multi-shot in-context learning~\citep{li2024long}, and repository-level code generation~\citep{bogomolov2024long}. These tasks typically require implicit reasoning steps that retrieve and aggregate information dispersed throughout extensive contexts before generating responses, posing significant challenges for contemporary long-context language models (LCLMs).

\begin{figure}
    \centering
    \includegraphics[width=1.0\linewidth]{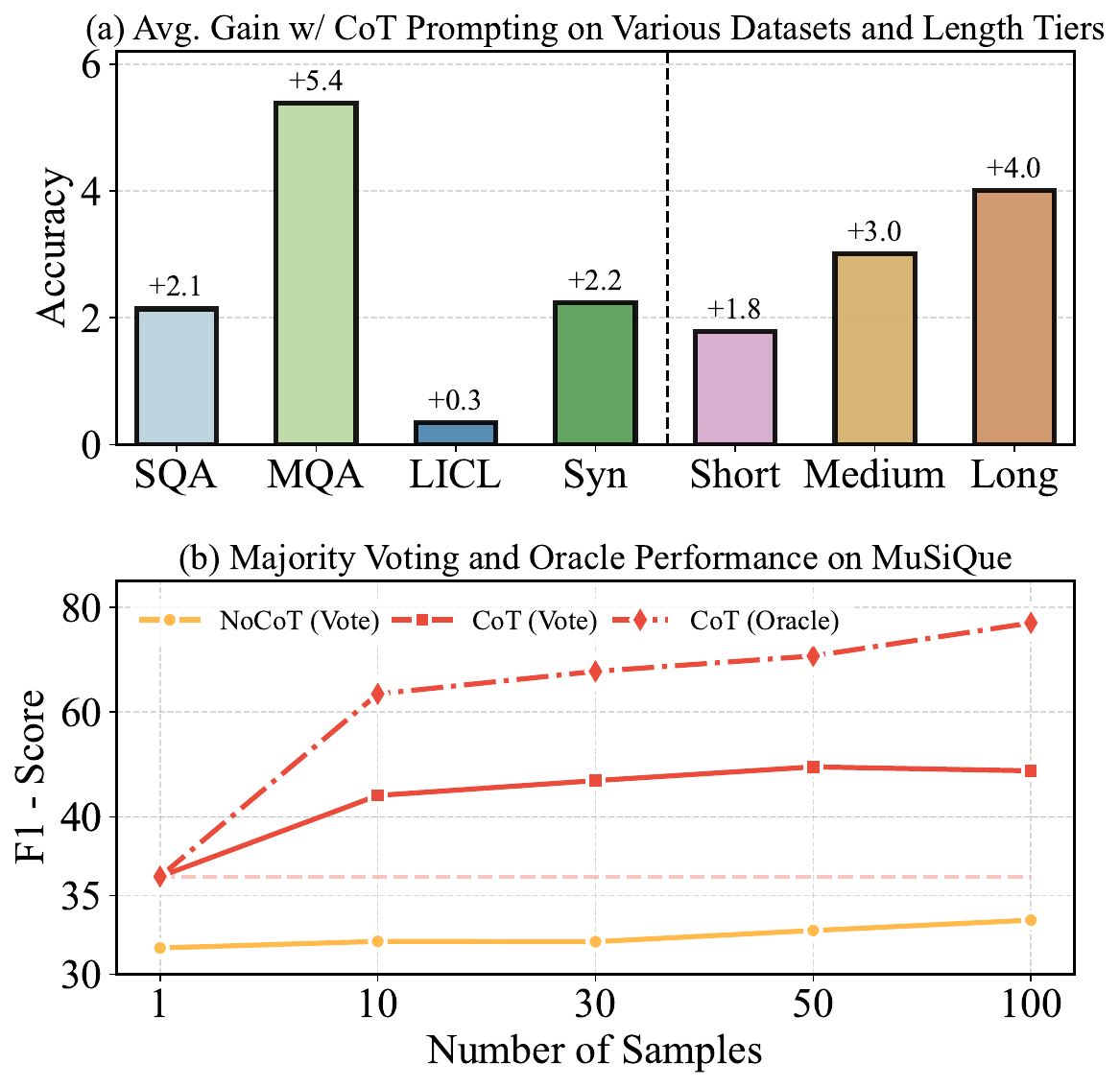}
    \caption{\textbf{(a)} Average gain w/ CoT prompting of open-source and proprietary models on long-context datasets of various domains and length tiers. \textit{SQA}, \textit{MQA}, \textit{LICL}, \textit{Syn} is short for \textit{Single-Document QA}, \textit{Multi-Document QA}, \textit{Long In-Context Learning}, and \textit{Synthetic} tasks, respectively. \textit{Short}, \textit{Medium}, \textit{Long} denotes different length tiers (<32k, 32-96k, >96k). Details see Section~\ref{sec:cot_effectiveness}. \textbf{(b)} Zero-shot majority voting results w.r.t. sampling rounds on MuSiQue, w/ and w/o CoT prompting.}
    \label{fig:intro_fig}
\end{figure}


As a means to elicit reasoning, Chain-of-Thought (CoT) prompting has demonstrated remarkable efficacy in enhancing multi-step reasoning tasks~\citep{wei2022chain,madaan2022text,sprague2024cot}. 
However, its effectiveness in long-context scenarios remains underexplored and only a few works have analyzed the use of CoT from limited perspectives, such as context length~\citep{modarressi2025nolima}, task difficulty~\citep{bai2024longbench2}, etc.
Building on previous efforts, we for the first time conduct a systematic investigation of CoT's effectiveness across a diverse set of long-context tasks of varying lengths and domains on both open-source and proprietary models of different scales. As illustrated in Figure~\ref{fig:intro_fig}(a), CoT's benefits generalize across most long-context scenarios, and amplify with increasing context length. Even more intriguingly, zero-shot majority voting results (Figure~\ref{fig:intro_fig}) on the long-context multi-hop reasoning task MuSiQue~\citep{trivedi2022musique} demonstrate that CoT substantially outperforms pure majority voting, and exhibits a steeper improvement trajectory as sampling rounds increase, but still significantly lags behind its oracle performance. This naturally leads to the following research question (\textbf{RQ}): \textit{How to enhance models' capability in long-context scenarios to generate high-quality reasoning paths~\footnote{In this work, we use the terms \textit{CoT} and \textit{reasoning path} interchangeably. They both refer to the models' reasoning thoughts including the final answer.} for improved performance?}

%

     
To address the question,  
we further investigate methods to train models in generating high-quality reasoning paths in long-context scenarios, aiming to enhance overall performance.
Given that real-world annotated data typically only contains outcome labels, we construct training data through a two-stage approach: first bootstrapping reasoning paths from the model itself, then identifying high-quality ones via a novel assessment protocol specifically designed for long-context scenarios. Our protocol evaluates both \textit{answer correctness} and \textit{process reliability}, with the latter strategically decomposed into \textit{source faithfulness} and \textit{intrinsic consistency} for efficient and accurate assessment. These self-sampling and quality assessment mechanisms together form a process-supervised framework that not only guides models toward correct answers but also teaches them appropriate reasoning patterns to reach these conclusions.

We train LLaMA3.1-8B~\citep{dubey2024llama} and Qwen2.5-7B~\citep{qwen2} on the MuSiQue dataset and comprehensively evaluate their performance across a diverse set of benchmarks, including: (1) MuSiQue, for assessing in-domain performance; (2) Selected QA tasks from LongBenchV1~\citep{bai2023longbench}, for measuring generalization capabilities on both single and multi-document question answering; and (3) Selected QA tasks from LongBenchV2~\citep{bai2024longbench2}, which feature longer contexts and more diverse domains. Compared to the baseline method of outcome supervision, our proposed process-supervised framework demonstrates superior performance on both the in-domain MuSiQue dataset (+13.6/+3.8 points for LLaMA/Qwen) and other QA tasks (+9.3/+8.1 points on average). These results validate the effectiveness of our approach across long-context scenarios spanning various domains and lengths.

Our contributions can be summarized as follows:
\begin{itemize}[leftmargin=*]
    \item We conduct, for the first time, a systematic examination to consolidate the effectiveness of CoT across most long-context scenarios of varying lengths and domains.
    \item We propose a long-context process-supervised framework, \ours, which comprises CoT sampling and a quality assessment protocol that efficiently ensures both answer correctness and process reliability of reasoning paths for long-context scenarios.
    \item Comprehensive experiments validate the superiority of \ours in both in-domain and generalization performance, and demonstrate the effectiveness of our quality assessment protocol.
\end{itemize}

\section{Related Work}
\label{sec:related_work}
\noindent \textbf{Long-Context Language Modeling}\quad
Context length is one of the most fundamental properties of language models, determining the total amount of information they can process at once. Long-context language modeling seeks to extend this capacity, pushing the boundaries of models' context windows. Research for context window extension primarily follows two directions. The first is data-driven strategies, which focus on constructing training data that exhibits long-range dependency patterns~\citep{gao2024prolong,fu2024data,prolong,dubey2024llama,xiong-etal-2024-effective,si2024selecting,wang2024bootstrap,Chen2024WhatAT,longalign,wu2024long}. The second is architecture-driven approaches, which enhance models' potential for processing long contexts by modifying components such as position encodings~\citep{Chen2023ExtendingCW,zhu2023pose,peng2024yarn,ding2024longrope} and attention mechanisms~\citep{an2024does,jin2024llm}. Building upon these advances, current LCLMs have demonstrated remarkable performance on simple long-context tasks, such as needle-in-haystack retrieval~\citep{needleinhaystack,hsieh2024ruler,zhu-etal-2024-longembed}. However, these models still struggle with more complex tasks, particularly long-context reasoning~\citep{kuratov2024babilong,li2024long,levy2024same}. Motivated by this challenge, our work investigates how to improve models' long-context reasoning capabilities by guiding them to generate high-quality Chains of Thought (CoT).


\noindent \textbf{Process Supervision}\quad 
Process supervision is a concept proposed to distinguish from outcome supervision~\citep{uesato2022solving}. In process supervision, models are not only guided toward correct answers but also taught appropriate reasoning patterns to reach these conclusions. A significant application of process supervision is in reinforcement learning~\citep{uesato2022solving,lightman2023let,luo2024improve}, where trained reward models for process evaluation, known as Process Reward Models (PRMs), are used to supervise the execution process of policy models. Numerous studies have demonstrated that process supervision outperforms outcome supervision in domains requiring multi-step reasoning, such as mathematics~\citep{wang2024math,lightman2023let}. Beyond the reinforcement learning paradigm, many works directly utilize high-quality CoT, obtained either through model's self-rejection sampling~\citep{zelikman2022star, singh2023beyond, zhang2024rest, hosseini2024vstar, pang2024iterative, wang2024self} or sampling from other models, to train models' reasoning process via supervised fine-tuning~\citep{QwQ,guo2025deepseek,team2025kimi}. Our approach falls into the latter category. Most closely related to our work is SEALong~\citep{li2024large}. Our method differs from it in two significant aspects: (1) While SEALong focuses on unsupervised scenarios with limited improvements, we achieve substantial in-domain and generalization gains by building upon outcome supervision in supervised settings. (2) We specifically design a CoT quality assessment protocol tailored for long-context scenarios to ensure both accuracy and reliability of generated CoTs, whereas SEALong primarily relies on outcomes to select CoTs.

\section{Efficacy of CoT in Long Context Tasks}
\label{sec:cot_effectiveness}

\noindent \textbf{Tasks}\quad To comprehensively evaluate the effectiveness of CoT in long context scenarios, we curate a diverse set of tasks encompassing both synthetic and real-world scenarios, covering a wide length range from 10k to 128k tokens. For synthetic scenarios, we select the single needle retrieval task~(S-NIAH) from RULER~\citep{hsieh2024ruler} and the multi-needle reasoning task with 3 supporting facts (MNR3) from BABILong~\citep{kuratov2024babilong}, which specifically focus on assessing models' retrieval and reasoning capabilities within extended contexts. Since synthetic datasets allow flexible length control, we sample an equal number of examples at lengths of \{16,32,64,128\}k for testing.
For real-world scenarios, we select three representative tasks from LongBench-V2~\citep{bai2024longbench2}: Single-Document Question-Answering~(SQA), Multi-Document Question-Answering~(MQA), and Long In-Context Learning~(LICL), covering various domains including Academic, Literary, Legal, Finance, etc. For each task type, we retain test samples with lengths less than 128k. Basic statistics of the curated evaluation dataset is presented in Table~\ref{tab:basic_statistics}.
\begin{figure*}[t]
    \centering
    \includegraphics[width=\textwidth]{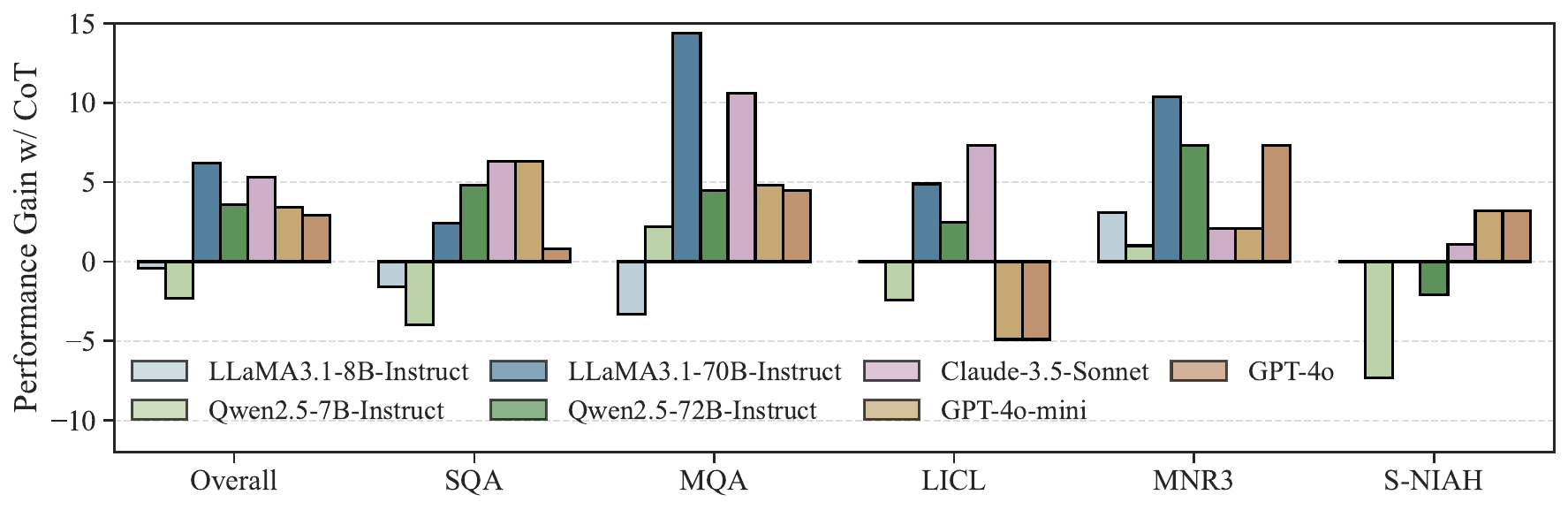}
    \caption{Performance gain of CoT for synthetic~(MNR3, S-NIAH) and real-world~(SQA, MQA, LICL) long context scenarios across all models. It is demonstrated that CoT particularly benefits proprietary and large-scale open-source models, and its effectiveness ranges across most long context scenarios, except for extremely easy retrieval tasks. }
    \label{fig:cot_gain}
\end{figure*}
In this section, we conduct a systematic investigation into the effectiveness of Chain-of-Thought in zero-shot long context scenarios. Below, we will first introduce our evaluation setting~(\S~\ref{sec:exp_setup}), and then present the detailed experimental results~(\S~\ref{sec:exp_sqa}).
\subsection{Evaluation Setup}
\label{sec:exp_setup}

\begin{table}[t]
    \centering
    \footnotesize
    \renewcommand{\arraystretch}{1.1}
    \setlength\tabcolsep{2.5pt}
    \begin{tabular}{l|ccc|cc|c}
    \toprule
    \multirow{2.5}{*}{\textbf{Tasks}} & \multicolumn{3}{c|}{\textbf{Real-World}} & \multicolumn{2}{c|}{\textbf{Synthetic}} & \multirow{2.5}{*}{\textbf{Total}}\\
    \cmidrule(r){2-4} \cmidrule(r){5-6}
    & \textbf{SQA} & \textbf{MQA} & \textbf{LICL} & \textbf{S-NIAH} & \textbf{MNR3} & \\
    \midrule
    \textbf{\#Data} & 126 & 90 & 41 & 96 & 96 & 449 \\
    \textbf{AvgLen} & 49k & 48k & 76k & 57k & 57k & 55k \\
    \textbf{Source} & \multicolumn{3}{c|}{LongBenchV2} & RULER & BABILong & - \\
    \bottomrule
    \end{tabular}
    \caption{Statistics of the curated evaluation dataset. \textit{AvgLen} is short for \textit{Average Length}, which is measured by number of tokens according to the tokenizer of LLaMA3.1~\citep{dubey2024llama}.}
    \label{tab:basic_statistics}
\end{table}

\noindent \textbf{Metrics}\quad To facilitate evaluation, we structure all tasks into a multiple-choice QA format to standardize the model's output format, which enables us to directly use choice accuracy as the evaluation metric. It is worth noting that the original tasks in RULER and BABILong are not presented in a multiple-choice format. Therefore, we create multiple-choice QA pairs by combining the current sample's ground truth with ground truth answers from other samples to form the candidate options.


\noindent \textbf{Models}\quad To comprehensively analyze the effectiveness of CoT in long context scenarios, we conduct experiments using both open-source and proprietary models of different scales and architectures, all supporting 128k context length. The tested models include LLaMA3.1-\{8B,70B\}-Instruct~\citep{dubey2024llama}, Qwen2.5-\{7B,72B\}-Instruct~\citep{yang2024qwen2}, Claude-3.5-Sonnet~\citep{claude3.5sonnet}, GPT-4o~\citep{GPT4o} and GPT-4o-mini~\citep{GPT4o}.

\subsection{Observations}
\label{sec:exp_sqa}

Figure~\ref{fig:cot_gain} presents inference-time performance gain of CoT for both synthetic and real-world long context scenarios across all models.


First, we observe that \textbf{CoT particularly benefits proprietary and large-scale open-source models (Observation 1)}, as evidenced by the improvement on LLaMA3.1-70B-Instruct, Qwen2.5-70B-Instruct, Claude-3.5-Sonnet, GPT-4o-mini. By contrast, small scale models including LLaMA3.1-8B-Instruct and Qwen2.5-7B-Instruct fail to yield consistent performance gains from CoT.
We hypothesize that this is due to smaller models' weaker overall capabilities, which may prevent them from generating high-quality reasoning paths to derive correct answers.

In addition, we find that \textbf{CoT is effective across most long context scenarios, except for extremely easy retrieval tasks (Observation 2)}. This is consolidated by the results that most models yield improvement on complex real-world scenarios (SQA, MQA, LICL) and synthetic scenarios that requires multi-hop reasoning over long context (MNR3), while showing no gain and even suffering loss on the easiest single needle retrieval task (S-NIAH). We attribute this to the fact that S-NIAH is already sufficiently simple for most LCLMs, typically achieving 100\% accuracy, and introducing CoT may actually create interference. Notably, while \citet{sprague2024cot} indicates that CoT is only effective for short-context tasks involving symbolic operations and reasoning, our observations complement their findings and demonstrate the importance of CoT prompting for long-context tasks. 


Further analysis explores how CoT benefits test samples of different lengths. We divide the test data into three length tiers: Short (<32k), Medium (32k-96k), and Long (>96k), with 199, 161, and 89 samples each. Table~\ref{tab:cot_length} presents the performance gains achieved with CoT across all large-scale models for each length interval. We observe that the Medium and Long groups generally show higher improvements with CoT compared to the Short group. This indicates that \textbf{CoT provides more benefits for longer tasks (Observation 3)}, possibly due to the increased need for implicit reasoning steps for information retrieval in longer contexts.


\begin{table}[t]
    \centering
    \footnotesize
    \renewcommand{\arraystretch}{1.1}
    \setlength\tabcolsep{6.5pt}
    \begin{tabular}{lccc}
    \toprule
    \textbf{Model} & \textbf{Short} & \textbf{Medium} & \textbf{Long} \\
    \midrule
    LLaMA3.1-70B-Instruct & +2.0 & +9.9 & +9.0 \\
    Qwen2.5-72B-Instruct & +1.0 & +5.0 & +6.7 \\
    Claude-3.5-Sonnet & +3.5 & +9.3 & +1.6 \\
    GPT-4o & -0.5 & +5.5 & +5.7 \\
    GPT-4o-mini & +4.0 & +1.3 & +6.2 \\ \midrule
    \textbf{Avg.} & +2.0 & +6.2 & +5.8 \\
    \bottomrule
    \end{tabular}
    \caption{Performance gain of CoT on test samples of different length tiers: Short (<32k), Medium (32k-96k), and Long (>96k). CoT generally provides more benefits for longer samples (Medium \& Long).}
    \label{tab:cot_length}
\end{table}



\begin{figure*}[htbp]
    \centering
    \includegraphics[width=1.0\textwidth]{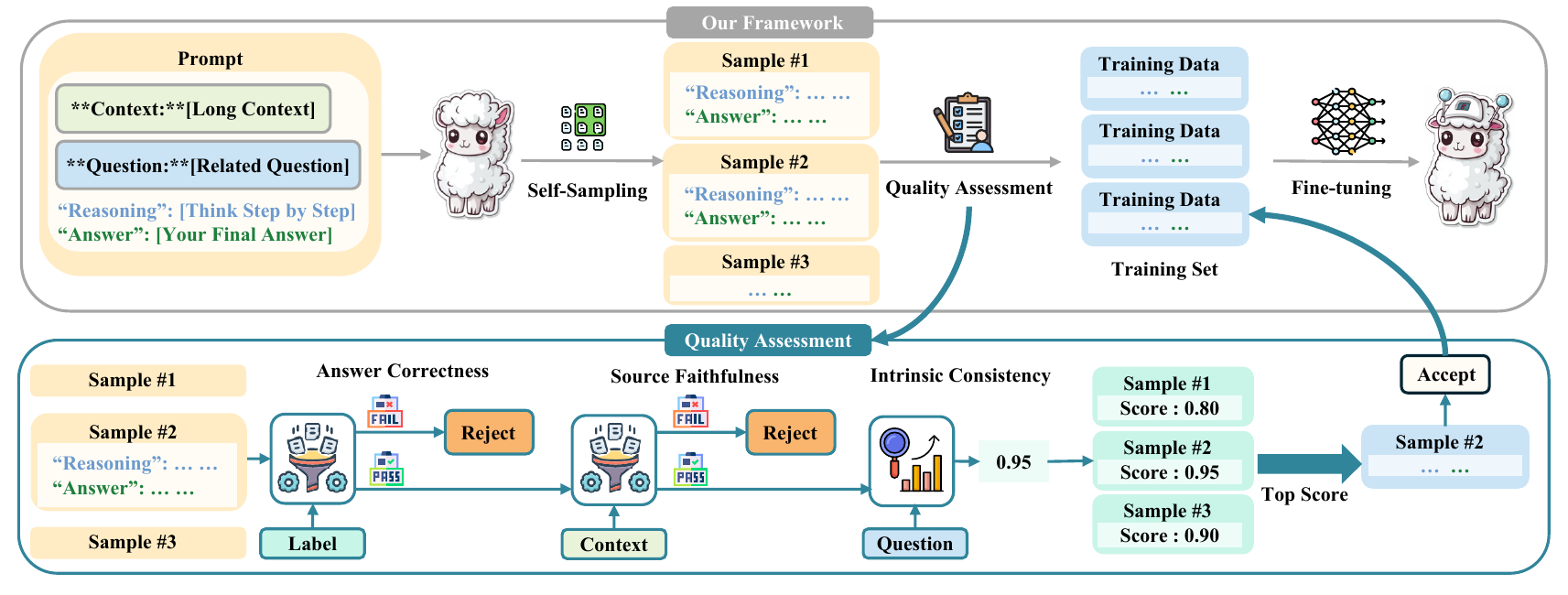}
    \caption{Our process-supervised framework \ours. We begin by sampling a diverse collection of $N$ reasoning paths from the model. A quality assessment procedure consisting of three criteria is then applied to these samples to select high-quality training samples, which are then used for supervised fine-tuning.}
    \label{fig:framework}
\end{figure*}

\section{Improving LCLMs via Self-Sampled Reasoning Paths}

Section~\ref{sec:cot_effectiveness} consolidates the substantial benefits of CoT in long-context scenarios. Encouraged by these findings, we further explore how to enhance models' ability to generate high-quality CoTs in long-context scenarios, thereby improving their reasoning capabilities over long contexts. This section details the proposed process-supervised framework \ours, which comprises a self-sampling phase to collect reasoning paths, a quality assessment phase specifically designed for long-context scenarios to indentify high-quality CoTs, and a training phase that performs supervised fine-tuning using the high-quality CoTs. The overall training process of \ours is illustrated in Figure~\ref{fig:framework}.

\subsection{Self-Sampling}
\label{sec:stagetwo}
First, we collect diverse reasoning paths through self-sampling on the training dataset. Specifically, for each training example, we generate a diverse collection of $N$ reasoning paths to encourage exploration of various reasoning strategies. During sampling, we employ a prompt~(See Appendix~\ref{app:cot_train}) that guides models through three key steps: (1) breaking down the question into manageable components, (2) identifying and citing relevant excerpts from the source text, and (3) deriving conclusions based on these excerpts. Crucially, we require models to explicitly cite source text by marking excerpts with `[Excerpt xxx]', where each excerpt must exactly match a portion of the original document. This strict matching requirement enables efficient and reliable assessment of source faithfulness in the subsequent quality evaluation phase~(Section~\ref{sec:cot_quality_assessment}).

\subsection{CoT Quality Assessment}
\label{sec:cot_quality_assessment}

\label{sec:filtering}
To identify high-quality CoT for supervised training, we evaluate CoT quality along two dimensions: \textit{answer correctness} and \textit{process reliability}, respectively ensuring the final answer aligns with the ground truth and the reasoning paths are logically coherent, concise, and well-supported.

\noindent \textbf{Answer Correctness (AC)}\quad 
First, we require CoT to arrive at correct answers, filtering out failed reasoning paths. Specifically, we compute the F1-score between the CoT-derived final answer and the ground truth, retaining only reasoning paths with F1-scores above a threshold $\delta$. This straightforward criterion ensures the basic effectiveness of the reasoning process.

\noindent \textbf{Process Reliability}\quad 
Beyond answer correctness, we further require the reasoning process itself to be reliable. Evaluating the reliability of reasoning processes is particularly challenging in long-context scenarios, as it requires referencing extensive input text. Even when using LLMs capable of handling long inputs, ensuring assessment accuracy remains difficult and computationally expensive. To address this challenge, we decompose process reliability into two components: \textit{source faithfulness} and \textit{intrinsic consistency}. The former ensures reasoning faithfulness to the source text and can be efficiently measured through simple string matching with our novel design, while the latter focuses on CoT's inherent quality and can be assessed by LLMs without requiring additional context. This decomposition enables efficient evaluation of reasoning processes in long-context scenarios.

\begin{itemize}[leftmargin=*]
    \item \textbf{Source Faithfulness}\quad Source Faithfulness requires reasoning paths to be grounded in the actual content present in the long context. This is crucial to prevent hallucination or the inclusion of irrelevant content that could contaminate the training data. By requiring models to provide relevant excerpts during sampling (Section 4.1), we can directly employ substring exact matching to verify citation faithfulness against the source text. This enables efficient assessment of CoT's information fidelity, allowing us to filter out reasoning paths that contain content inconsistent with the original text.
    \item \textbf{Intrinsic Consistency}\quad After ensuring source faithfulness to the input text, we further examine the quality of CoT itself, which we term as Intrinsic Consistency. Specifically, a high-quality CoT should demonstrate logical coherence (appropriate question breakdown, logical use of information, and sound reasoning chain), completeness (primary reliance on retrieved information rather than model's internal knowledge), and conciseness (avoiding irrelevant or excessive details). Given the complexity of evaluating these dimensions, we employ LLM scoring with prompts detailed in Appendix~\ref{sec:prompts}.

\end{itemize}

Finally, for each retained example, we select the CoT with the highest intrinsic consistency score for inclusion in the training data. In this way, we construct a high-quality self-sampled dataset for further model fine-tuning.




    

    

    

    

\subsection{Supervised Fine-tuning}

After collecting high-quality self-sampled reasoning paths, we enhance the model's ability to generate high-quality CoT through supervised fine-tuning, thereby comprehensively improving its performance on long-context tasks. Specifically, we minimize the expected negative log-likelihood:
\[
\mathcal{L}_{\theta} = \mathbb{E}_{(x,y)\sim\mathcal{D}}[-\log\mathcal{M}_{\theta}(y|x)]
\]
where \(\mathcal{D}\) denotes the training dataset, and \(\mathcal{M}_{\theta}(y|x)\) denotes the conditional likelihood of CoT \(y\) under the model parameterized by \(\theta\).

\begin{table*}[htbp]
    \centering
    \footnotesize
    \renewcommand{\arraystretch}{1.1}
    \setlength\tabcolsep{4pt}
    \begin{tabular}{lcccccccc}
        \toprule
            \multirow{2.5}{*}{\textbf{Model}} & \multirow{2.5}{*}{\textbf{MuSiQue}} & \multicolumn{4}{c}{\textbf{QAs-LBV1}} & \multicolumn{2}{c}{\textbf{QAs-LBV2}} & \multirow{2.5}{*}{\textbf{Avg.}} \\
            \cmidrule(lr){3-6} \cmidrule(lr){7-8}
            &  & \textbf{HPQA} & \textbf{MFQA} & \textbf{Qasper} & \textbf{2WMQA} & \textbf{SQA} & \textbf{MQA} & \\
            \midrule
            \rowcolor{gray!15} LLaMA-3.1-8B-Instruct & 45.3 & 57.3 & 53.8 & 42.9 & 64.0 & 29.3 & 32.2 & 46.4 \\
            \textit{LLaMA-3.1-8B-Base} & 12.6 & 21.2 & 29.9 & 13.4 & 19.9 & 1.2 & 1.7 & 14.3 \\
            \quad w/ Outcome Supervision & 47.0 & 50.0 & 44.4 & 32.1 & 37.1 & 17.1 & 14.8 & 34.7 \\
            \quad w/ \ours & 
            60.6{\scriptsize (\textbf{+13.6})}{\cellcolor{darkgreen}} & 
            57.9{\scriptsize (\textbf{+7.9})}{\cellcolor{mediumgreen}} & 
            53.8{\scriptsize (\textbf{+9.4})}{\cellcolor{mediumgreen}} & 
            36.0{\scriptsize (\textbf{+3.9})}{\cellcolor{mediumgreen}} & 
            50.5{\scriptsize (\textbf{+13.4})}{\cellcolor{darkgreen}} & 
            28.1{\scriptsize (\textbf{+11.0})}{\cellcolor{darkgreen}} & 
            31.3{\scriptsize (\textbf{+16.5})}{\cellcolor{darkgreen}} & 
            44.0{\scriptsize (\textbf{+9.3})}{\cellcolor{mediumgreen}} \\
            \midrule
            \rowcolor{gray!15} Qwen-2.5-7B-Instruct & 39.4 & 57.5 & 48.7 & 43.0 & 54.2 & 34.2 & 33.0 & 44.3 \\
            \textit{Qwen-2.5-7B-Base} & 23.8 & 43.8 & 46.2 & 29.9 & 28.2 & 30.5 & 32.2 & 33.5 \\ 
            \quad w/ Outcome Supervision & 49.2 & 58.1 & 43.2 & 29.7 & 41.2 & 20.7 & 17.4 & 37.1 \\
            \quad w/ \ours & 53.0{\scriptsize (\textbf{+3.8})}{\cellcolor{lightgreen}} & 
            57.0{\scriptsize (\textbf{-1.1})}{\cellcolor{lightred}} & 
            45.6{\scriptsize (\textbf{+2.4})}{\cellcolor{lightgreen}} & 
            38.4{\scriptsize (\textbf{+8.7})}{\cellcolor{mediumgreen}} & 
            58.1{\scriptsize (\textbf{+16.9})}{\cellcolor{darkgreen}} & 
            30.5{\scriptsize (\textbf{+9.8})}{\cellcolor{mediumgreen}} & 
            33.9{\scriptsize (\textbf{+16.5})}{\cellcolor{darkgreen}} & 
            45.2{\scriptsize (\textbf{+8.1})}{\cellcolor{mediumgreen}} \\
            \midrule
            GPT-4o-mini & 46.3 & 56.1 & 50.2 & 38.7 & 64.0 & 34.2 & 34.2 & 46.2 \\
            GPT-4o & 55.8 & 65.8 & 54.8 & 45.4 & 74.8 & 46.0 & 47.2 & 55.7 \\
        \bottomrule
    \end{tabular}
    \caption{Performance comparison of different models on the selected QA tasks from LongBenchV1~(\textit{QAs-LBV1}) and the longer and more domain-diverse QAs from LongBenchV2~(\textit{QAs-LBV2}). \textit{HPQA}, \textit{MFQA}, \textit{2WMQA}, \textit{SQA}, \textit{MQA} is short for HotpotQA, MultiFieldQA, 2WikiMultihopQA, Single-Document QA, Multi-Document QA,respectively. Numbers in parentheses show changes from outcome supervision baseline, (improvements in \textcolor{deepdarkgreen}{green}, degradations in \textcolor{red}{red}). \ours not only excels on in-domain tasks (MuSiQue) but also exhibits strong generalization capabilities, improving across long-context scenarios of varying domains, and context lengths.}
    \label{tab:main_results} 
\end{table*}

\section{Experiments}

\subsection{Experimental Setups}
\noindent \textbf{Training Data}\quad We conduct our self-training experiments on MuSiQue~\citep{trivedi2022musique}, a challenging multi-hop QA task that requires models to integrate information from multiple Wikipedia documents to answer questions. To better align with modern long-context scenarios, we extend the dataset's context length from its original average of less than 4k tokens to a range of 10-16k tokens, following the methodology proposed in SEALong~\citep{li2024large}.

\noindent \textbf{Candidate Models}\quad We select LLaMA-3.1-8B-Base~\cite{dubey2024llama} and Qwen-2.5-7B-Base~\cite{qwen2} as the candidate models for CoT self-training. Notably, we opt for base versions rather than instruction-tuned variants, as the latter typically undergo sophisticated post-training procedures that may include CoT generation optimizations, which could compromise the reliability of our experimental results. To equip the base models with basic instruction-following capabilities, we perform a simple warmup process (details provided in \textit{Implementation Details}), before proceeding to the self-sampling stage.

\noindent \textbf{Evaluation Setup}\quad To comprehensively analyze the effectiveness of our proposed self-training framework, we evaluate the trained models across a diverse set of datasets, including: 
(1) MuSiQue, for assessing in-domain performance; (2) Selected QA tasks from LongBenchV1~\citep{bai2023longbench}, including Qasper~\citep{dasigi2021dataset}, HotpotQA~\citep{yang2018hotpotqa}, MultiFieldQA-En~\citep{bai2023longbench}, and 2WikiMultihopQA~\citep{ho2020constructing}, for measuring generalization capabilities on both single and multi-document question answering; and (3) Selected QA tasks from LongBenchV2~\citep{bai2024longbench2}, which feature longer contexts and more diverse domains. We employ F1-score and multiple-choice accuracy as evaluation metrics, depending on the format in which each task is structured. We also include the results from for LLaMA3.1-8B-Instruct, Qwen2.5-7B-Instruct, GPT-4o, GPT-4o-mini for reference.


\noindent \textbf{Implementation Details}\quad Our training dataset for MuSiQue comprises 3,300 examples in total. For the warmup stage, we first select 300 examples and obtain their reasoning paths from the instruction-tuned versions of LLaMA-3.1-8B and Qwen-2.5-7B, then fine-tune their corresponding base models on these CoT examples for 20 steps, using a constant learning rate of $1e^{-5}$ and a global batch size of 32. In the self-training stage, we use the warmed-up models to generate reasoning paths for the remaining 3,000 examples with a temperature of 0.7. Specifically, we sample 30 reasoning paths per example and apply our proposed filtering strategy (Section~\ref{sec:filtering}) to construct the training set. We set the answer consistency threshold to 1.0, which filters out approximately 1,000 examples. The remaining examples are used to fine-tune the warmed-up models for 2 epochs, with a constant learning rate of $5e^{-6}$ and a global batch size of 32. For the baseline method using outcome supervision, we directly fine-tune the warmed-up models on the complete dataset of 3,000 examples for 2 epochs, using identical hyperparameters. For fair comparison, we report the best performance achieved across the two epochs. All training procedures are conducted on 8 A100 GPUs using LLaMA-Factory~\citep{zheng2024llamafactory,360-llama-factory}.

\subsection{Main Results}
Table~\ref{tab:main_results} presents the performance of our models across the evaluated datasets.

\noindent \textbf{CoT Supervision Beats Outcome Supervision}\quad 
First, we observe that our proposed method significantly outperforms the baseline method of outcome supervision on MuSiQue for both \textit{LLaMA-3.1-8B-Base} and \textit{Qwen-2.5-7B-Base}. The improvement is particularly pronounced for the \textit{LLaMA-3.1-8B-Base} model, demonstrating a substantial performance gain of +13.6 points. Similarly, in the model's majority voting performance evaluation, our method also achieves a significant improvement (See Appendix~\ref{sec:vote}). This validates the overall effectiveness of CoT supervision, which not only guides models toward correct answers but also teaches them appropriate reasoning patterns to reach these conclusions.

\begin{table}[t]
    \centering
    \footnotesize
    \renewcommand{\arraystretch}{1.1}
    \setlength\tabcolsep{2.5pt}
    \begin{tabular}{lcccc}
        \toprule
            \textbf{Assess Criteria} & \textbf{MuSiQue} & \textbf{QAs-LBV1} & \textbf{QAs-LBV2} & \textbf{Avg.} \\
            \midrule 
            \multicolumn{5}{c}{\textit{LLaMA-3.1-8B}} \\ \midrule
            AC check only  & 54.0 & 46.7 & 22.9 & 40.9 \\
            + SF & 54.8 & 46.3 & 25.1 & 41.4 \\
            + SF + IC & \textbf{60.6} & \textbf{47.0} & \textbf{29.7} & \textbf{44.0} \\
            \midrule
            \multicolumn{5}{c}{\textit{Qwen-2.5-7B}} \\ \midrule
            AC check only & 49.8 & 49.7 & 29.2 & 43.9 \\
            + SF & 50.9 & \textbf{50.1} & 31.8 & 45.0 \\
            + SF + IC & \textbf{53.0} & 49.8 & \textbf{32.2} & \textbf{45.2}  \\

        \bottomrule
    \end{tabular}
    \caption{Ablation experiments on our proposed CoT quality assessment protocol. \textit{AC}, \textit{SF}, and \textit{IC} are short for \textit{Answer Correctness}, \textit{Source Faithfulness}, and \textit{Intrinsic Consistency}, respectively. Other notations follow Table~\ref{tab:main_results}. The results show that each assessment criterion contributes positively to the model's performance, and the model achieves the highest performance when all three criteria are applied together.}
    \label{tab:ablation_results} 
    \end{table}

\noindent \textbf{Superior Generalizability}\quad 
Furthermore, Table~\ref{tab:main_results} compares the generalization performance of our method against the baseline across various tasks. Results show that our method consistently achieves superior performance, both on other multi-document QA tasks from LongBenchV1 and on tasks from LongBenchV2 featuring longer contexts and more diverse domains. Specifically, we observe average performance gains of +9.3 and +8.1 points for LLaMA and Qwen models respectively, reaching performance levels comparable to GPT-4o-mini. These results demonstrate that our method not only excels on in-domain tasks but also exhibits strong generalization capabilities, bringing improvements across long-context scenarios of varying formats, domains, and context lengths.

\subsection{Ablation Study}
\noindent \textbf{Efficacy of Our Quality Assessment Protocol}\quad
We further examine the effectiveness of our assessment protocol in selecting high-quality CoTs as training data. Since maintaining answer correctness is a fundamental requirement and de facto standard in common practice, its effectiveness is self-evident. We thus focus on examining the criteria for process reliability: source faithfulness and intrinsic consistency.
As shown in Table~\ref{tab:ablation_results}, incorporating each assessment criterion brings positive benefits, with the highest performance achieved when all three criteria work together. For the LLaMA-3.1-8B model, compared to using answer consistency check alone, adding source faithfulness check leads to an average gain of +0.5 points, while further incorporating intrinsic consistency scoring brings an additional +2.6 points. Similarly, for the Qwen-2.5-7B model, applying all three assessment criteria improves the average score by +1.3 points over using answer consistency check alone. This consistent improvement across both models underscores the necessity of each filtering component in ensuring the quality of self-sampled reasoning paths.

\section{Analysis}
\begin{figure}[t]
\centering
\includegraphics[width=\linewidth]{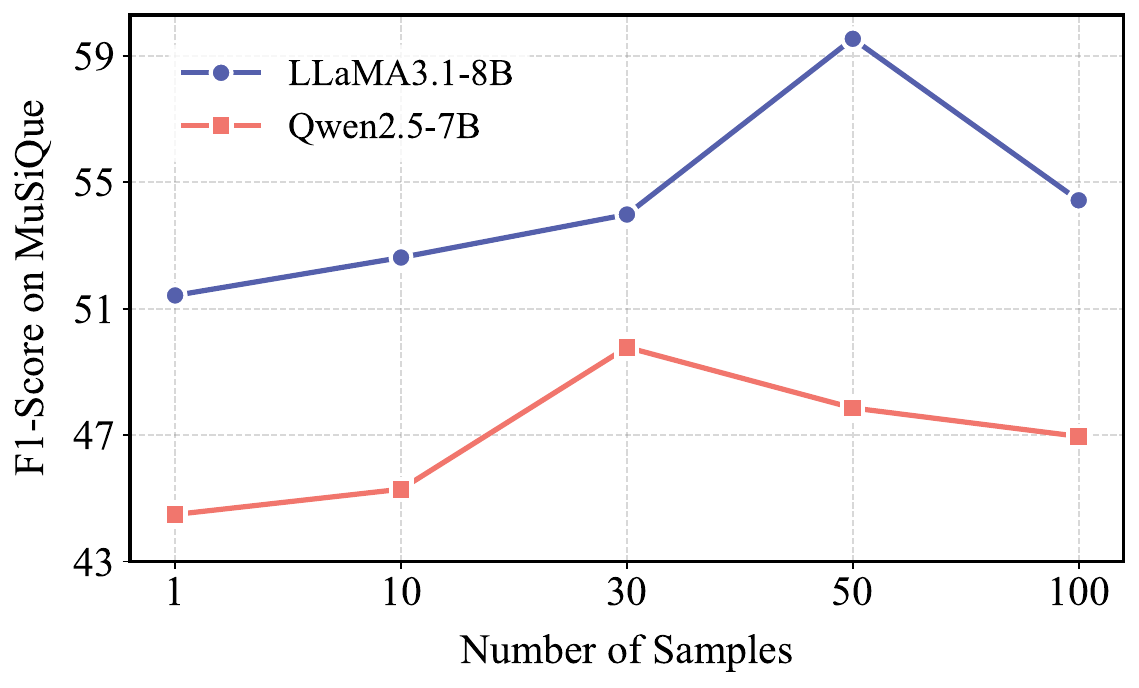}
\caption{Impact of sampling size on model performance on MuSiQue. }
\label{fig:impact_of_the_number_of_sample_examples}
\end{figure}
\noindent \textbf{Impact of Sampling Size}\quad
First, we analyze how the number of sampled candidate CoTs per training example affects model performance. As expected, a larger sampling size provides more diverse candidates, potentially enabling the selection of higher-quality CoTs. Figure~\ref{fig:impact_of_the_number_of_sample_examples} illustrates the impact of sampling size on model performance on the test set of MuSiQue. Interestingly, we observe that model performance first increases and then decreases with the increasing sampling size. For Qwen2.5-7B, performance peaks when the sampling number N reaches 30, while for Llama3.1-8B, the optimal performance is achieved at N=50. We hypothesize that this phenomenon might be due to the increased difficulty in maintaining consistent quality assessment with larger candidate pools.

\paragraph{Impact of CoT Source Models}

\begin{figure}[t]
\centering
\includegraphics[width=1.0\linewidth]{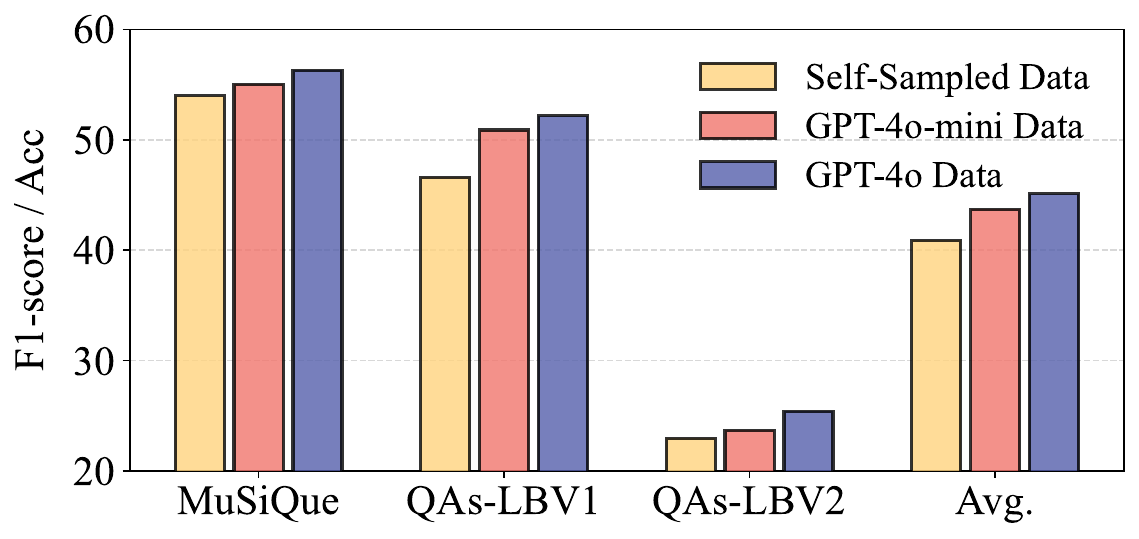}
\captionsetup{skip=2pt}
\caption{The impact of CoTs of different quality on model performance. Notations align with Table~\ref{tab:main_results}.}
\label{fig:impact_of_the_quality_of_CoT}
\end{figure}

Beyond self-sampling CoTs from the base model, we investigate the effectiveness of training with CoTs directly sampled from more capable models (GPT-4o-mini and GPT-4o), as shown in Figure~\ref{fig:impact_of_the_quality_of_CoT}. We observe that GPT-4o-sampled CoTs consistently achieve the best performance across all scenarios: the in-domain MuSiQue task, other QA tasks from LongBenchV1, and more diverse, longer-context QA tasks from LongBenchV2, followed by GPT-4o-mini and then self-sampled CoTs. Combined with our ablation results in Table~\ref{tab:ablation_results}, these findings demonstrate that improving CoT quality, whether through careful selection from self-sampled candidates or direct sampling from stronger models, effectively enhances model capabilities in long-context scenarios.

\section{Conclusion}
In this study, we propose \ours, a process-supervised framework which enhances the long-context reasoning capabilities of LLM by guiding them to generate high-quality reasoning paths. By conducting experiments on various long-context benchmarks, we demonstrate that our framework significantly improves model performance on both in-domain and out-domain tasks, highlighting the generalization capabilities of our method. Our experiments underscore the critical role of process-supervised reasoning paths in enhancing long-context language models, enabling them to navigate and reason over extensive textual information with greater accuracy and reliability. By reducing the reliance on human annotations and leveraging self-sampled data, our approach offers a scalable solution for enhancing LLMs' reasoning abilities.
\section*{Limitations}
Our work is still limited in some aspects. Although we have conducted experiments on Llama-3.1-8B and Qwen-2.5-7B to verify the validity of our proposed framework, we lack experiments on larger models (e.g., 32B and 70B parameters) due to limited computational resources. Similarly, due to GPU memory constraints, we can only fine-tune our model on sequences up to 16K tokens in length. We believe that fine-tuning on longer sequences would lead to better performance on long-text tasks. In future work, we plan to address these limitations and further explore the effectiveness of our framework on larger models and longer sequences.

\section*{Acknowledgement}

\bibliography{bib/custom}

\begin{thebibliography}{55}
\providecommand{\natexlab}[1]{#1}

\bibitem[{An et~al.(2024)An, Zhang, Zhong, Li, Gong, Luo, Xu, and Kong}]{an2024does}
Chenxin An, Jun Zhang, Ming Zhong, Lei Li, Shansan Gong, Yao Luo, Jingjing Xu, and Lingpeng Kong. 2024.
\newblock Why does the effective context length of llms fall short?
\newblock \emph{arXiv preprint arXiv:2410.18745}.

\bibitem[{Bai et~al.(2024{\natexlab{a}})Bai, Lv, Zhang, He, Qi, Hou, Tang, Dong, and Li}]{longalign}
Yushi Bai, Xin Lv, Jiajie Zhang, Yuze He, Ji~Qi, Lei Hou, Jie Tang, Yuxiao Dong, and Juanzi Li. 2024{\natexlab{a}}.
\newblock \href {https://doi.org/10.18653/v1/2024.findings-emnlp.74} {{L}ong{A}lign: A recipe for long context alignment of large language models}.
\newblock In \emph{Findings of the Association for Computational Linguistics: EMNLP 2024}, pages 1376--1395, Miami, Florida, USA. Association for Computational Linguistics.

\bibitem[{Bai et~al.(2023)Bai, Lv, Zhang, Lyu, Tang, Huang, Du, Liu, Zeng, Hou et~al.}]{bai2023longbench}
Yushi Bai, Xin Lv, Jiajie Zhang, Hongchang Lyu, Jiankai Tang, Zhidian Huang, Zhengxiao Du, Xiao Liu, Aohan Zeng, Lei Hou, et~al. 2023.
\newblock Longbench: A bilingual, multitask benchmark for long context understanding.
\newblock \emph{arXiv preprint arXiv:2308.14508}.

\bibitem[{Bai et~al.(2024{\natexlab{b}})Bai, Tu, Zhang, Peng, Wang, Lv, Cao, Xu, Hou, Dong, Tang, and Li}]{bai2024longbench2}
Yushi Bai, Shangqing Tu, Jiajie Zhang, Hao Peng, Xiaozhi Wang, Xin Lv, Shulin Cao, Jiazheng Xu, Lei Hou, Yuxiao Dong, Jie Tang, and Juanzi Li. 2024{\natexlab{b}}.
\newblock Longbench v2: Towards deeper understanding and reasoning on realistic long-context multitasks.
\newblock \emph{arXiv preprint arXiv:2412.15204}.

\bibitem[{Bogomolov et~al.(2024)Bogomolov, Eliseeva, Galimzyanov, Glukhov, Shapkin, Tigina, Golubev, Kovrigin, van Deursen, Izadi et~al.}]{bogomolov2024long}
Egor Bogomolov, Aleksandra Eliseeva, Timur Galimzyanov, Evgeniy Glukhov, Anton Shapkin, Maria Tigina, Yaroslav Golubev, Alexander Kovrigin, Arie van Deursen, Maliheh Izadi, et~al. 2024.
\newblock Long code arena: a set of benchmarks for long-context code models.
\newblock \emph{arXiv preprint arXiv:2406.11612}.

\bibitem[{Brown et~al.(2020)Brown, Mann, Ryder, Subbiah, Kaplan, Dhariwal, Neelakantan, Shyam, Sastry, Askell et~al.}]{brown2020language}
Tom Brown, Benjamin Mann, Nick Ryder, Melanie Subbiah, Jared~D Kaplan, Prafulla Dhariwal, Arvind Neelakantan, Pranav Shyam, Girish Sastry, Amanda Askell, et~al. 2020.
\newblock Language models are few-shot learners.
\newblock \emph{Advances in neural information processing systems}, 33:1877--1901.

\bibitem[{Chen et~al.(2024{\natexlab{a}})Chen, Liu, He, Zheng, Sun, Li, Luo, and Yang}]{prolong}
Longze Chen, Ziqiang Liu, Wanwei He, Yinhe Zheng, Hao Sun, Yunshui Li, Run Luo, and Min Yang. 2024{\natexlab{a}}.
\newblock \href {https://doi.org/10.18653/v1/2024.acl-long.447} {Long context is not long at all: A prospector of long-dependency data for large language models}.
\newblock In \emph{Proceedings of the 62nd Annual Meeting of the Association for Computational Linguistics (Volume 1: Long Papers)}, pages 8222--8234, Bangkok, Thailand. Association for Computational Linguistics.

\bibitem[{Chen et~al.(2023)Chen, Wong, Chen, and Tian}]{Chen2023ExtendingCW}
Shouyuan Chen, Sherman Wong, Liangjian Chen, and Yuandong Tian. 2023.
\newblock Extending context window of large language models via positional interpolation.
\newblock \emph{ArXiv}, abs/2306.15595.

\bibitem[{Chen et~al.(2024{\natexlab{b}})Chen, Chen, Qin, Guo, Lv, Zou, Che, Yan, Chen, and Lin}]{Chen2024WhatAT}
Zhi Chen, Qiguang Chen, Libo Qin, Qipeng Guo, Haijun Lv, Yicheng Zou, Wanxiang Che, Hang Yan, Kai Chen, and Dahua Lin. 2024{\natexlab{b}}.
\newblock What are the essential factors in crafting effective long context multi-hop instruction datasets? insights and best practices.
\newblock \emph{ArXiv}, abs/2409.01893.

\bibitem[{Claude(2024)}]{claude3.5sonnet}
Team Claude. 2024.
\newblock Claude 3.5 sonnet.
\newblock \url{https://www.anthropic.com/news/claude-3-5-sonnet}.

\bibitem[{Dasigi et~al.(2021)Dasigi, Lo, Beltagy, Cohan, Smith, and Gardner}]{dasigi2021dataset}
Pradeep Dasigi, Kyle Lo, Iz~Beltagy, Arman Cohan, Noah~A Smith, and Matt Gardner. 2021.
\newblock A dataset of information-seeking questions and answers anchored in research papers.
\newblock In \emph{Proceedings of the 2021 Conference of the North American Chapter of the Association for Computational Linguistics: Human Language Technologies}, pages 4599--4610.

\bibitem[{Ding et~al.(2024)Ding, Zhang, Zhang, Xu, Shang, Xu, Yang, and Yang}]{ding2024longrope}
Yiran Ding, Li~Lyna Zhang, Chengruidong Zhang, Yuanyuan Xu, Ning Shang, Jiahang Xu, Fan Yang, and Mao Yang. 2024.
\newblock Longrope: Extending llm context window beyond 2 million tokens.
\newblock In \emph{Forty-first International Conference on Machine Learning}.

\bibitem[{Dubey et~al.(2024)Dubey, Jauhri, Pandey, Kadian, Al-Dahle, Letman, Mathur, Schelten, Yang, Fan et~al.}]{dubey2024llama}
Abhimanyu Dubey, Abhinav Jauhri, Abhinav Pandey, Abhishek Kadian, Ahmad Al-Dahle, Aiesha Letman, Akhil Mathur, Alan Schelten, Amy Yang, Angela Fan, et~al. 2024.
\newblock The llama 3 herd of models.
\newblock \emph{arXiv preprint arXiv:2407.21783}.

\bibitem[{Fu et~al.(2024)Fu, Panda, Niu, Yue, Hajishirzi, Kim, and Peng}]{fu2024data}
Yao Fu, Rameswar Panda, Xinyao Niu, Xiang Yue, Hannaneh Hajishirzi, Yoon Kim, and Hao Peng. 2024.
\newblock Data engineering for scaling language models to 128k context.
\newblock In \emph{Forty-first International Conference on Machine Learning}.

\bibitem[{Gao et~al.(2024)Gao, Wettig, Yen, and Chen}]{gao2024prolong}
Tianyu Gao, Alexander Wettig, Howard Yen, and Danqi Chen. 2024.
\newblock How to train long-context language models (effectively).
\newblock \emph{arXiv preprint arXiv:2410.02660}.

\bibitem[{GPT4o(2024)}]{GPT4o}
Team GPT4o. 2024.
\newblock Hello gpt-4o.
\newblock \url{https://openai.com/index/hello-gpt-4o/}.

\bibitem[{Guo et~al.(2025)Guo, Yang, Zhang, Song, Zhang, Xu, Zhu, Ma, Wang, Bi et~al.}]{guo2025deepseek}
Daya Guo, Dejian Yang, Haowei Zhang, Junxiao Song, Ruoyu Zhang, Runxin Xu, Qihao Zhu, Shirong Ma, Peiyi Wang, Xiao Bi, et~al. 2025.
\newblock Deepseek-r1: Incentivizing reasoning capability in llms via reinforcement learning.
\newblock \emph{arXiv preprint arXiv:2501.12948}.

\bibitem[{Haosheng~Zou and Zhang(2024)}]{360-llama-factory}
Shousheng~Jia Haosheng~Zou, Xiaowei~Lv and Xiangzheng Zhang. 2024.
\newblock \href {https://github.com/Qihoo360/360-LLaMA-Factory} {360-llama-factory}.

\bibitem[{Ho et~al.(2020)Ho, Nguyen, Sugawara, and Aizawa}]{ho2020constructing}
Xanh Ho, Anh-Khoa~Duong Nguyen, Saku Sugawara, and Akiko Aizawa. 2020.
\newblock Constructing a multi-hop qa dataset for comprehensive evaluation of reasoning steps.
\newblock \emph{arXiv preprint arXiv:2011.01060}.

\bibitem[{Hosseini et~al.(2024)Hosseini, Yuan, Malkin, Courville, Sordoni, and Agarwal}]{hosseini2024vstar}
Arian Hosseini, Xingdi Yuan, Nikolay Malkin, Aaron Courville, Alessandro Sordoni, and Rishabh Agarwal. 2024.
\newblock V-{ST}ar: Training verifiers for self-taught reasoners.
\newblock In \emph{First Conference on Language Modeling}.

\bibitem[{Hsieh et~al.(2024)Hsieh, Sun, Kriman, Acharya, Rekesh, Jia, Zhang, and Ginsburg}]{hsieh2024ruler}
Cheng-Ping Hsieh, Simeng Sun, Samuel Kriman, Shantanu Acharya, Dima Rekesh, Fei Jia, Yang Zhang, and Boris Ginsburg. 2024.
\newblock Ruler: What's the real context size of your long-context language models?
\newblock \emph{arXiv preprint arXiv:2404.06654}.

\bibitem[{Jin et~al.(2024)Jin, Han, Yang, Jiang, Liu, Chang, Chen, and Hu}]{jin2024llm}
Hongye Jin, Xiaotian Han, Jingfeng Yang, Zhimeng Jiang, Zirui Liu, Chia-Yuan Chang, Huiyuan Chen, and Xia Hu. 2024.
\newblock Llm maybe longlm: Selfextend llm context window without tuning.
\newblock In \emph{Forty-first International Conference on Machine Learning}.

\bibitem[{Kamradt(2023)}]{needleinhaystack}
Greg Kamradt. 2023.
\newblock Needle in a haystack - pressure testing llms.
\newblock \url{https://github.com/gkamradt/LLMTest_NeedleInAHaystack}.

\bibitem[{Kuratov et~al.(2024)Kuratov, Bulatov, Anokhin, Rodkin, Sorokin, Sorokin, and Burtsev}]{kuratov2024babilong}
Yuri Kuratov, Aydar Bulatov, Petr Anokhin, Ivan Rodkin, Dmitry Sorokin, Artyom Sorokin, and Mikhail Burtsev. 2024.
\newblock Babilong: Testing the limits of llms with long context reasoning-in-a-haystack.
\newblock \emph{arXiv preprint arXiv:2406.10149}.

\bibitem[{Levy et~al.(2024)Levy, Jacoby, and Goldberg}]{levy2024same}
Mosh Levy, Alon Jacoby, and Yoav Goldberg. 2024.
\newblock Same task, more tokens: the impact of input length on the reasoning performance of large language models.
\newblock \emph{arXiv preprint arXiv:2402.14848}.

\bibitem[{Li et~al.(2024{\natexlab{a}})Li, Yang, Cheng, Liu, Yu, Yang, and Lam}]{li2024large}
Siheng Li, Cheng Yang, Zesen Cheng, Lemao Liu, Mo~Yu, Yujiu Yang, and Wai Lam. 2024{\natexlab{a}}.
\newblock Large language models can self-improve in long-context reasoning.
\newblock \emph{arXiv preprint arXiv:2411.08147}.

\bibitem[{Li et~al.(2024{\natexlab{b}})Li, Zhang, Do, Yue, and Chen}]{li2024long}
Tianle Li, Ge~Zhang, Quy~Duc Do, Xiang Yue, and Wenhu Chen. 2024{\natexlab{b}}.
\newblock Long-context llms struggle with long in-context learning.
\newblock \emph{arXiv preprint arXiv:2404.02060}.

\bibitem[{Lightman et~al.(2023)Lightman, Kosaraju, Burda, Edwards, Baker, Lee, Leike, Schulman, Sutskever, and Cobbe}]{lightman2023let}
Hunter Lightman, Vineet Kosaraju, Yura Burda, Harri Edwards, Bowen Baker, Teddy Lee, Jan Leike, John Schulman, Ilya Sutskever, and Karl Cobbe. 2023.
\newblock Let's verify step by step.
\newblock \emph{arXiv preprint arXiv:2305.20050}.

\bibitem[{Luo et~al.(2024)Luo, Liu, Liu, Phatale, Lara, Li, Shu, Zhu, Meng, Sun et~al.}]{luo2024improve}
Liangchen Luo, Yinxiao Liu, Rosanne Liu, Samrat Phatale, Harsh Lara, Yunxuan Li, Lei Shu, Yun Zhu, Lei Meng, Jiao Sun, et~al. 2024.
\newblock Improve mathematical reasoning in language models by automated process supervision.
\newblock \emph{arXiv preprint arXiv:2406.06592}.

\bibitem[{Madaan and Yazdanbakhsh(2022)}]{madaan2022text}
Aman Madaan and Amir Yazdanbakhsh. 2022.
\newblock Text and patterns: For effective chain of thought, it takes two to tango.
\newblock \emph{arXiv preprint arXiv:2209.07686}.

\bibitem[{Modarressi et~al.(2025)Modarressi, Deilamsalehy, Dernoncourt, Bui, Rossi, Yoon, and Sch{\"u}tze}]{modarressi2025nolima}
Ali Modarressi, Hanieh Deilamsalehy, Franck Dernoncourt, Trung Bui, Ryan~A Rossi, Seunghyun Yoon, and Hinrich Sch{\"u}tze. 2025.
\newblock Nolima: Long-context evaluation beyond literal matching.
\newblock \emph{arXiv preprint arXiv:2502.05167}.

\bibitem[{Pang et~al.(2024)Pang, Yuan, He, Cho, Sukhbaatar, and Weston}]{pang2024iterative}
Richard~Yuanzhe Pang, Weizhe Yuan, He~He, Kyunghyun Cho, Sainbayar Sukhbaatar, and Jason~E Weston. 2024.
\newblock Iterative reasoning preference optimization.
\newblock In \emph{The Thirty-eighth Annual Conference on Neural Information Processing Systems}.

\bibitem[{Peng et~al.(2024)Peng, Quesnelle, Fan, and Shippole}]{peng2024yarn}
Bowen Peng, Jeffrey Quesnelle, Honglu Fan, and Enrico Shippole. 2024.
\newblock Yarn: Efficient context window extension of large language models.
\newblock In \emph{The Twelfth International Conference on Learning Representations}.

\bibitem[{QwQ(2024)}]{QwQ}
Team QwQ. 2024.
\newblock Qwq.
\newblock \url{https://qwenlm.github.io/blog/qwq-32b-preview/}.

\bibitem[{Si et~al.(2024)Si, Zhao, Chen, Li, Luo, Lv, An, Qi, Chang, and Sun}]{si2024selecting}
Shuzheng Si, Haozhe Zhao, Gang Chen, Yunshui Li, Kangyang Luo, Chuancheng Lv, Kaikai An, Fanchao Qi, Baobao Chang, and Maosong Sun. 2024.
\newblock Selecting influential samples for long context alignment via homologous models’ guidance and contextual awareness measurement.

\bibitem[{Singh et~al.(2023)Singh, Co-Reyes, Agarwal, Anand, Patil, Garcia, Liu, Harrison, Lee, Xu et~al.}]{singh2023beyond}
Avi Singh, John~D Co-Reyes, Rishabh Agarwal, Ankesh Anand, Piyush Patil, Xavier Garcia, Peter~J Liu, James Harrison, Jaehoon Lee, Kelvin Xu, et~al. 2023.
\newblock Beyond human data: Scaling self-training for problem-solving with language models.
\newblock \emph{arXiv preprint arXiv:2312.06585}.

\bibitem[{Sprague et~al.(2024)Sprague, Yin, Rodriguez, Jiang, Wadhwa, Singhal, Zhao, Ye, Mahowald, and Durrett}]{sprague2024cot}
Zayne Sprague, Fangcong Yin, Juan~Diego Rodriguez, Dongwei Jiang, Manya Wadhwa, Prasann Singhal, Xinyu Zhao, Xi~Ye, Kyle Mahowald, and Greg Durrett. 2024.
\newblock To cot or not to cot? chain-of-thought helps mainly on math and symbolic reasoning.
\newblock \emph{arXiv preprint arXiv:2409.12183}.

\bibitem[{Team et~al.(2025)Team, Du, Gao, Xing, Jiang, Chen, Li, Xiao, Du, Liao et~al.}]{team2025kimi}
Kimi Team, Angang Du, Bofei Gao, Bowei Xing, Changjiu Jiang, Cheng Chen, Cheng Li, Chenjun Xiao, Chenzhuang Du, Chonghua Liao, et~al. 2025.
\newblock Kimi k1. 5: Scaling reinforcement learning with llms.
\newblock \emph{arXiv preprint arXiv:2501.12599}.

\bibitem[{Trivedi et~al.(2022)Trivedi, Balasubramanian, Khot, and Sabharwal}]{trivedi2022musique}
Harsh Trivedi, Niranjan Balasubramanian, Tushar Khot, and Ashish Sabharwal. 2022.
\newblock Musique: Multihop questions via single-hop question composition.
\newblock \emph{Transactions of the Association for Computational Linguistics}, 10:539--554.

\bibitem[{Uesato et~al.(2022)Uesato, Kushman, Kumar, Song, Siegel, Wang, Creswell, Irving, and Higgins}]{uesato2022solving}
Jonathan Uesato, Nate Kushman, Ramana Kumar, Francis Song, Noah Siegel, Lisa Wang, Antonia Creswell, Geoffrey Irving, and Irina Higgins. 2022.
\newblock Solving math word problems with process-and outcome-based feedback.
\newblock \emph{arXiv preprint arXiv:2211.14275}.

\bibitem[{Wang et~al.(2024{\natexlab{a}})Wang, Yang, Zhang, Huang, and Wei}]{wang2024bootstrap}
Liang Wang, Nan Yang, Xingxing Zhang, Xiaolong Huang, and Furu Wei. 2024{\natexlab{a}}.
\newblock Bootstrap your own context length.
\newblock \emph{arXiv preprint arXiv:2412.18860}.

\bibitem[{Wang et~al.(2024{\natexlab{b}})Wang, Li, Shao, Xu, Dai, Li, Chen, Wu, and Sui}]{wang2024math}
Peiyi Wang, Lei Li, Zhihong Shao, Runxin Xu, Damai Dai, Yifei Li, Deli Chen, Yu~Wu, and Zhifang Sui. 2024{\natexlab{b}}.
\newblock Math-shepherd: Verify and reinforce llms step-by-step without human annotations.
\newblock In \emph{Proceedings of the 62nd Annual Meeting of the Association for Computational Linguistics (Volume 1: Long Papers)}, pages 9426--9439.

\bibitem[{Wang et~al.(2024{\natexlab{c}})Wang, Li, and Lu}]{wang2024self}
Tianduo Wang, Shichen Li, and Wei Lu. 2024{\natexlab{c}}.
\newblock Self-training with direct preference optimization improves chain-of-thought reasoning.
\newblock In \emph{Proceedings of the 62nd Annual Meeting of the Association for Computational Linguistics (Volume 1: Long Papers)}, pages 11917--11928.

\bibitem[{Wei et~al.(2022)Wei, Wang, Schuurmans, Bosma, Xia, Chi, Le, Zhou et~al.}]{wei2022chain}
Jason Wei, Xuezhi Wang, Dale Schuurmans, Maarten Bosma, Fei Xia, Ed~Chi, Quoc~V Le, Denny Zhou, et~al. 2022.
\newblock Chain-of-thought prompting elicits reasoning in large language models.
\newblock \emph{Advances in neural information processing systems}, 35:24824--24837.

\bibitem[{Wu et~al.(2024)Wu, Wang, Fu, Yue, Zhu, and Li}]{wu2024long}
Wenhao Wu, Yizhong Wang, Yao Fu, Xiang Yue, Dawei Zhu, and Sujian Li. 2024.
\newblock Long context alignment with short instructions and synthesized positions.
\newblock \emph{arXiv preprint arXiv:2405.03939}.

\bibitem[{Xiong et~al.(2024)Xiong, Liu, Molybog, Zhang, Bhargava, Hou, Martin, Rungta, Sankararaman, Oguz, Khabsa, Fang, Mehdad, Narang, Malik, Fan, Bhosale, Edunov, Lewis, Wang, and Ma}]{xiong-etal-2024-effective}
Wenhan Xiong, Jingyu Liu, Igor Molybog, Hejia Zhang, Prajjwal Bhargava, Rui Hou, Louis Martin, Rashi Rungta, Karthik~Abinav Sankararaman, Barlas Oguz, Madian Khabsa, Han Fang, Yashar Mehdad, Sharan Narang, Kshitiz Malik, Angela Fan, Shruti Bhosale, Sergey Edunov, Mike Lewis, Sinong Wang, and Hao Ma. 2024.
\newblock \href {https://doi.org/10.18653/v1/2024.naacl-long.260} {Effective long-context scaling of foundation models}.
\newblock In \emph{Proceedings of the 2024 Conference of the North American Chapter of the Association for Computational Linguistics: Human Language Technologies (Volume 1: Long Papers)}, pages 4643--4663, Mexico City, Mexico. Association for Computational Linguistics.

\bibitem[{Yang et~al.(2024{\natexlab{a}})Yang, Yang, Hui, Zheng, Yu, Zhou, Li, Li, Liu, Huang, Dong, Wei, Lin, Tang, Wang, Yang, Tu, Zhang, Ma, Xu, Zhou, Bai, He, Lin, Dang, Lu, Chen, Yang, Li, Xue, Ni, Zhang, Wang, Peng, Men, Gao, Lin, Wang, Bai, Tan, Zhu, Li, Liu, Ge, Deng, Zhou, Ren, Zhang, Wei, Ren, Fan, Yao, Zhang, Wan, Chu, Liu, Cui, Zhang, and Fan}]{qwen2}
An~Yang, Baosong Yang, Binyuan Hui, Bo~Zheng, Bowen Yu, Chang Zhou, Chengpeng Li, Chengyuan Li, Dayiheng Liu, Fei Huang, Guanting Dong, Haoran Wei, Huan Lin, Jialong Tang, Jialin Wang, Jian Yang, Jianhong Tu, Jianwei Zhang, Jianxin Ma, Jin Xu, Jingren Zhou, Jinze Bai, Jinzheng He, Junyang Lin, Kai Dang, Keming Lu, Keqin Chen, Kexin Yang, Mei Li, Mingfeng Xue, Na~Ni, Pei Zhang, Peng Wang, Ru~Peng, Rui Men, Ruize Gao, Runji Lin, Shijie Wang, Shuai Bai, Sinan Tan, Tianhang Zhu, Tianhao Li, Tianyu Liu, Wenbin Ge, Xiaodong Deng, Xiaohuan Zhou, Xingzhang Ren, Xinyu Zhang, Xipin Wei, Xuancheng Ren, Yang Fan, Yang Yao, Yichang Zhang, Yu~Wan, Yunfei Chu, Yuqiong Liu, Zeyu Cui, Zhenru Zhang, and Zhihao Fan. 2024{\natexlab{a}}.
\newblock Qwen2 technical report.
\newblock \emph{arXiv preprint arXiv:2407.10671}.

\bibitem[{Yang et~al.(2024{\natexlab{b}})Yang, Yang, Zhang, Hui, Zheng, Yu, Li, Liu, Huang, Wei et~al.}]{yang2024qwen2}
An~Yang, Baosong Yang, Beichen Zhang, Binyuan Hui, Bo~Zheng, Bowen Yu, Chengyuan Li, Dayiheng Liu, Fei Huang, Haoran Wei, et~al. 2024{\natexlab{b}}.
\newblock Qwen2. 5 technical report.
\newblock \emph{arXiv preprint arXiv:2412.15115}.

\bibitem[{Yang et~al.(2018)Yang, Qi, Zhang, Bengio, Cohen, Salakhutdinov, and Manning}]{yang2018hotpotqa}
Zhilin Yang, Peng Qi, Saizheng Zhang, Yoshua Bengio, William~W Cohen, Ruslan Salakhutdinov, and Christopher~D Manning. 2018.
\newblock Hotpotqa: A dataset for diverse, explainable multi-hop question answering.
\newblock \emph{arXiv preprint arXiv:1809.09600}.

\bibitem[{Zelikman et~al.(2022)Zelikman, Wu, Mu, and Goodman}]{zelikman2022star}
Eric Zelikman, Yuhuai Wu, Jesse Mu, and Noah Goodman. 2022.
\newblock {ST}ar: Bootstrapping reasoning with reasoning.
\newblock In \emph{Advances in Neural Information Processing Systems}.

\bibitem[{Zhang et~al.(2024)Zhang, Zhoubian, Yue, Dong, and Tang}]{zhang2024rest}
Dan Zhang, Sining Zhoubian, Yisong Yue, Yuxiao Dong, and Jie Tang. 2024.
\newblock Rest-mcts*: Llm self-training via process reward guided tree search.
\newblock \emph{arXiv preprint arXiv:2406.03816}.

\bibitem[{Zheng et~al.(2024)Zheng, Zhang, Zhang, Ye, Luo, Feng, and Ma}]{zheng2024llamafactory}
Yaowei Zheng, Richong Zhang, Junhao Zhang, Yanhan Ye, Zheyan Luo, Zhangchi Feng, and Yongqiang Ma. 2024.
\newblock \href {http://arxiv.org/abs/2403.13372} {Llamafactory: Unified efficient fine-tuning of 100+ language models}.
\newblock In \emph{Proceedings of the 62nd Annual Meeting of the Association for Computational Linguistics (Volume 3: System Demonstrations)}, Bangkok, Thailand. Association for Computational Linguistics.

\bibitem[{Zhong et~al.(2021)Zhong, Yin, Yu, Zaidi, Mutuma, Jha, Hassan, Celikyilmaz, Liu, Qiu et~al.}]{zhong2021qmsum}
Ming Zhong, Da~Yin, Tao Yu, Ahmad Zaidi, Mutethia Mutuma, Rahul Jha, Ahmed Hassan, Asli Celikyilmaz, Yang Liu, Xipeng Qiu, et~al. 2021.
\newblock Qmsum: A new benchmark for query-based multi-domain meeting summarization.
\newblock In \emph{Proceedings of the 2021 Conference of the North American Chapter of the Association for Computational Linguistics: Human Language Technologies}, pages 5905--5921.

\bibitem[{Zhu et~al.(2024)Zhu, Wang, Yang, Song, Wu, Wei, and Li}]{zhu-etal-2024-longembed}
Dawei Zhu, Liang Wang, Nan Yang, Yifan Song, Wenhao Wu, Furu Wei, and Sujian Li. 2024.
\newblock \href {https://doi.org/10.18653/v1/2024.emnlp-main.47} {{L}ong{E}mbed: Extending embedding models for long context retrieval}.
\newblock In \emph{Proceedings of the 2024 Conference on Empirical Methods in Natural Language Processing}, pages 802--816, Miami, Florida, USA. Association for Computational Linguistics.

\bibitem[{Zhu et~al.(2023)Zhu, Yang, Wang, Song, Wu, Wei, and Li}]{zhu2023pose}
Dawei Zhu, Nan Yang, Liang Wang, Yifan Song, Wenhao Wu, Furu Wei, and Sujian Li. 2023.
\newblock Pose: Efficient context window extension of llms via positional skip-wise training.
\newblock In \emph{The Twelfth International Conference on Learning Representations}.

\end{thebibliography}
\bibliographystyle{bib/acl_natbib}

\newpage
\appendix
\onecolumn

\section{Prompts}
\label{sec:prompts}

\begin{promptbox}[Prompt for Self-Sampling and Fine-tuning with CoT]{promptgreen}
\label{app:cot_train}
You are given a long document such as a story, meeting script, a news article, etc, and a question. Your task is to answer the question based on the information provided in the document. You should follow the instructions below to provide an accurate reasoning path, as well as a concise answer to the question: \\

**Instructions:**\\
Step 1. **Reasoning:** First retrieve all relevant information, then deduce the correct answer. Begin by carefully reading the provided context. Identify and extract all relevant information that is directly related to the question. Be succinct and only extract the most important excerpts that will help you answer the question. Finally, deduce the correct answer based on the retrieved information.\\
Step 2. **Answer:** Using the information you have retrieved, and your deduction, answer the question as concisely as you can, using a single phrase or sentence if possible. Ensure that your answer should be brief and to the point.\\
Step 3. **Format Your Response:** Present your response in JSON format, comprising two components: "reasoning" and "answer". The "reasoning" section should detail your thought process, including the breakdown of the question, the relevant excerpts (indicated by [Excerpt xxx] at the start), and the derived conclusion. Ensure that each excerpt is an exact match to the original document. Limit the number of excerpts to a maximum of 10. The "answer" part should contain your final answer to the question, as concise and to the point as possible.\\

Illustrative Examples:\\
Example \#1:\\
**Context:** [... Saltram is living with the Mulvilles at Wimbledon ... He is not working or producing anything ... He is idle and dependent on others ...]\\
**Question:** What is Saltram's living situation?\\
**Response:**\\
\{\{\\
     "reasoning": "Let me first retrieve relevant excerpts from the document, then answer the question. The question asks about Saltram's living situation. In the document, I can first locate that [Excerpt 1] `Saltram is living with the Mulvilles at Wimbledon`. Additionally, it is mentioned that [Excerpt 2] `He is not working or producing anything` and [Excerpt 3] `He is idle and dependent on others`. From these excerpts, I can deduce that Saltram is a guest in the home of the Mulvilles.",\\
    "answer": "He is a guest in the home of the Mulvilles."\\
\}\}\\

Example \#2:\\
**Context:** [... The Collegian is the bi-weekly official student publication of Houston Baptist University in Houston, Texas ... Houston Baptist University, affiliated with the Baptist General Convention of Texas, offers bachelor's and graduate degrees. It was founded in 1960 ...]\\
**Question:** When was the institute that owned The Collegian founded?\\
**Response:**\\
\{\{\\
    "reasoning": "Let me first retrieve relevant excerpts from the document, then answer the question. The question asks about the founding date of the institute that owned The Collegian. In the document, I can first locate that [Excerpt 1] `The Collegian is the bi-weekly official student publication of Houston Baptist University in Houston, Texas`, so I need to look for information about Houston Baptist University. I find that [Excerpt 2] `Houston Baptist University was founded in 1960`. Therefore, the institute that owned The Collegian was founded in 1960.",\\
    "answer": "1960"\\
\}\}\\

Now, based on the context provided below, answer the question as concisely as you can, using a single phrase or sentence if possible.\\
**Context:** \{context\}\\
**Question:** \{question\}\\
**Response:**\\
\end{promptbox}

\clearpage

\begin{promptbox}[Prompt for LLM Scoring]{promptgreen}
[Question]\\
\{question\}\\

[The Start of Assistant's Reasoning Path]\\
\{reasoning\} \\ \
[The End of Assistant's Reasoning Path]\\

[System]\\
We would like to request your feedback on the quality of the reasoning process in the given response. 
The model receives a long text input and a complex question. Its task is to retrieve relevant information from the long text (marked as [Excerpt xxx] and enclosed in ``) based on the question's requirements and provide the correct answer. Above, we have provided both the question and the model's reasoning process. While the model's final answer is correct, we need you to evaluate whether its reasoning process is sound.\\

Please assess the model's reasoning process based on the following aspects:\\

1. Logical Coherence:\\
- The model should break down the question appropriately\\
- The use of retrieved information should follow logical patterns\\
- The chain of reasoning from retrieved information to the final answer should be sound\\

2. Completeness:\\
- The reasoning process should primarily rely on information retrieved from the text ([Excerpts xxx])\\
- The model should not heavily depend on its own knowledge base\\

3. Conciseness:\\
- Only information relevant to answering the question should be retrieved\\
- The model should avoid listing excessive or irrelevant information\\

Please rate whether this reasoning path is suitable for the question. The assistant receives an overall score on a scale of 1 to 100, where a higher score indicates better overall performance.\\
Please note that if the assistant's reasoning process fully meets the above criteria, its overall rating should be full marks (100).\\
Please first provide a comprehensive explanation of your evaluation, avoiding any potential bias.\\
Then, output a line indicating the score of the Assistant.\\

PLEASE OUTPUT WITH THE FOLLOWING FORMAT, WHERE THE SCORE IS ON A SCALE OF 1 TO 100 BY STRICTLY FOLLOWING THIS FORMAT: "[[score]]", FOR EXAMPLE "Rating: [[100]]":\\
<start output>\\
Evaluation evidence: your evaluation explanation here, no more than 100 words\\
Rating: [[score]]\\
<end output>\\

Now, start your evaluation:
\end{promptbox}

\clearpage

\twocolumn
\section{Majority Voting Performance}
\label{sec:vote}

\begin{figure}[h]
\centering
\includegraphics[width=1.0\linewidth]{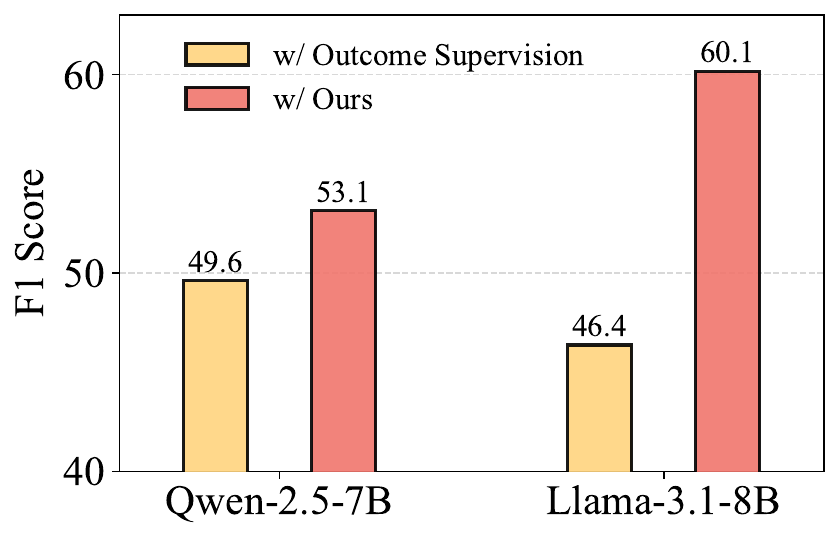}
\captionsetup{skip=2pt}
\caption{Majority voting results on MuSiQue, w/ outcome supervision and w/ our proposed method. The models align with Table~\ref{tab:main_results}.}
\label{fig:majority_vote}
\end{figure}

The majority voting results of Llama-3.1-8B and Qwen-2.5-7B are shown in Figure~\ref{fig:majority_vote}. We can see that for both models, our method shows a significant improvement compared to outcome supervision in the majority voting setting. This is consistent with the phenomena we observed in Table~\ref{tab:main_results}.

\clearpage

\end{document}